\documentclass[sigconf]{acmart}
\setcopyright{none}
\renewcommand\footnotetextcopyrightpermission[1]{} 
\makeatletter
\renewcommand\@formatdoi[1]{\ignorespaces}
\makeatother
\usepackage{comment}
\usepackage{algorithm}
\usepackage{algorithmic}
\AtBeginDocument{%
  \providecommand\BibTeX{{%
    \normalfont B\kern-0.5em{\scshape i\kern-0.25em b}\kern-0.8em\TeX}}}

\acmConference[CIKM '19]{CIKM '19: ACM International Conference on Information and Knowledge Management }{November 03-07, 2019}{Beijing, China}

\begin{document}

\title{SUM: Suboptimal Unitary Multi-task Learning Framework for Spatiotemporal Data Prediction}

\author{Qichen Li}
\email{liqichen@whu.edu.cn}
\affiliation{%
  \institution{Wuhan University}
}

\author{Jiaxin Pei}
\affiliation{%
  \institution{Wuhan University}
}
\email{pedropei@whu.edu.cn}

\author{Jianding Zhang}
\affiliation{%
  \institution{Wuhan University}
}
\email{zhangjianding@whu.edu.cn}

\author{Bo Han}
\authornote{Contact Author}
\affiliation{%
 \institution{Wuhan University}
}
\email{bhan@whu.edu.cn}


\begin{abstract}

In this paper, we propose a two-step suboptimal unitary method (SUM) for spatial-temporal data prediction. Inspired by the recent success of pretraining models in natural language processing, SUM is composed of a \textbf{global training} step, which searches for a global pattern by optimizing the general parameters with gradient descent under geographical constraints and a \textbf{task-specific training} step in which the pre-trained global model is further fine-tuned on a single task. We apply SUM to MUSCAT, a classical multi-task model for spatial-temporal prediction and have achieved better performance than MUSCAT and other baselines. Moreover, SUM allows MUSCAT to be directly transferred to unseen tasks, and shows better generalization ability than other coKriging methods.

\end{abstract}

\maketitle

\section{Introduction}
Spatio-temporal data, also known as spatial data streams, is a type of data describing information in different locations in a streaming manner(e.g. temperature information in contiguous cities over a period of time).  Due to the ubiquity of spatio-temporal data, how to computationally model it and further predict the future trend has attracted extensive attention of the data mining community. Considering the homogeneous nature of spatio-temperal data, current state-of-the-art prediction algorithms mainly follow the multi-task learning fashion, where predicting temporal information in different locations are viewed as tasks which differ in details but share similar temporal patterns. In this case, the majority of the multi-task researches focus on the extraction of the invariants, or the essentials of information, across space or time. For example, \cite{yu2015accelerated} devised low-rank tensor learning strategy to combine the parameters of each location into a tensor, which could reduce its rank while maintaining the prediction accuracy. \cite{kolda2009tensor} used tensor decomposition to extract essential information in spatio-temporal data streams.


While achieving promising results over several datasets, many existing approaches like low rank tensor learning lack generalizability - they cannot handle new locations whose data are not provided in the training set. Besides, some methods are specially designed for this setting (usually named coKriging), they either require geographic information from the new tasks\cite{bahadori2014fast}\cite{liu2018distribution}, which indicates their incapability on geographically unknown places, or have relatively low accuracy and high computational complexity\cite{bonilla2008multi}.


To address these problems, we advocate a new perspective to extract essential information apart from low-rank tensor learning. Inspired by the meta-learning methods\cite{finn2017model} and recent success of pretraining methods in NLP\cite{Devlin2018BERTPO}, we comprehend that each task has its own optimal solution but shares a common suboptimal solution due to their high correlation. Therefore, we come up with Suboptimal Unitary Multi-task(SUM) Learning framework for spatio-temporal data, which could serve as a structure to improve the performance of almost all multi-task learning models using gradient descent. Our first step is a pre-training process which seeks for this common suboptimal set of parameters, and our second step is a fine-tuning process which makes small adjustments considering each different task. We apply SUM to MUSCAT, a classical multi-task learning model for spatial data prediction. Experimenting over a public dataset, the enhanced model demonstrates its better prediction performance and generalizability than existing methods.

Our contributions are threefold:
\begin{itemize}

\item SUM offers a new perspective for spatial-temporal prediction that the relation between different tasks could be exploited by the proposed two step learning framework.

\item SUM is extremely flexible, thus we can apply it to many current models regardless of whether they are linear or not. 

\item SUM enables traditional prediction models to coKriging or make predictions even when geographical information is unavailable.
\end{itemize}




\section{Proposed Framework}
In this section, we first introduce the preliminaries for our proposed spatial-temporal prediction and then illustrate the two-step learning process of SUM.


\subsection{Preliminaries}
\subsubsection{Tensor, Matrix, and Vector}
We use calligraphy font for tensors(e.g. $\mathcal{X,Y}$)  , bold uppercase letters for matrices (e.g. $\boldsymbol{A,B}$), and bold lowercase letters for vectors(e.g. $\boldsymbol{x,y}$).

\subsubsection{Data and Variables}
Let \(\mathcal{D} =(\mathcal{X,Y})\) denote the give data.
We use $S,N$ and $M$ to denote the number of locations (tasks), predictor variables and response variables in the training set, respectively. For each location, the length of the given time series is denoted by $T$. Thus we have the predictor variable tensor \(\mathcal{X} \in \mathbf{R}^{N\times T \times S}\) and the response variable tensor \(Y \in \mathbf{R} ^ { M\times T \times S }\). For the task $i$, the corresponding learnable parameter is \(\theta _ { i }\). Besides, we use $vec(\cdot)$ to represent the vectorization operation. This operator means sequencing one or more multi-dimensional tensors into a vector in a fixed order.

\subsubsection{Problem Definition}
Formally, given predictor variables \(\mathcal{X} \in \mathbf{R}^{N\times T \times S}\), response variables \(Y \in \mathbf{R} ^ { M\times T \times S }\) and  a specific model $f(\cdot;\theta)$ which maps the predictor variable to the corresponding response vector. The goal is to find a set of parameter $\theta$ which satisfies 
\[\theta_* = \mathop{\arg\min}_{\theta}\sum_{i=1}^S L(f(\mathcal{X}_{:,:,i},\theta_i);\mathcal{Y}_{:,:,i})\eqno{(1)}\]

where the loss function L is usually defined as:
\[L(\hat { \mathcal{Y}}, \mathcal{Y}) = \|\hat { \mathcal{Y}}- \mathcal{Y}\|_{2}^{2}\eqno{(2)}\]

\begin{figure}
    \centering
    \includegraphics[width=8cm]{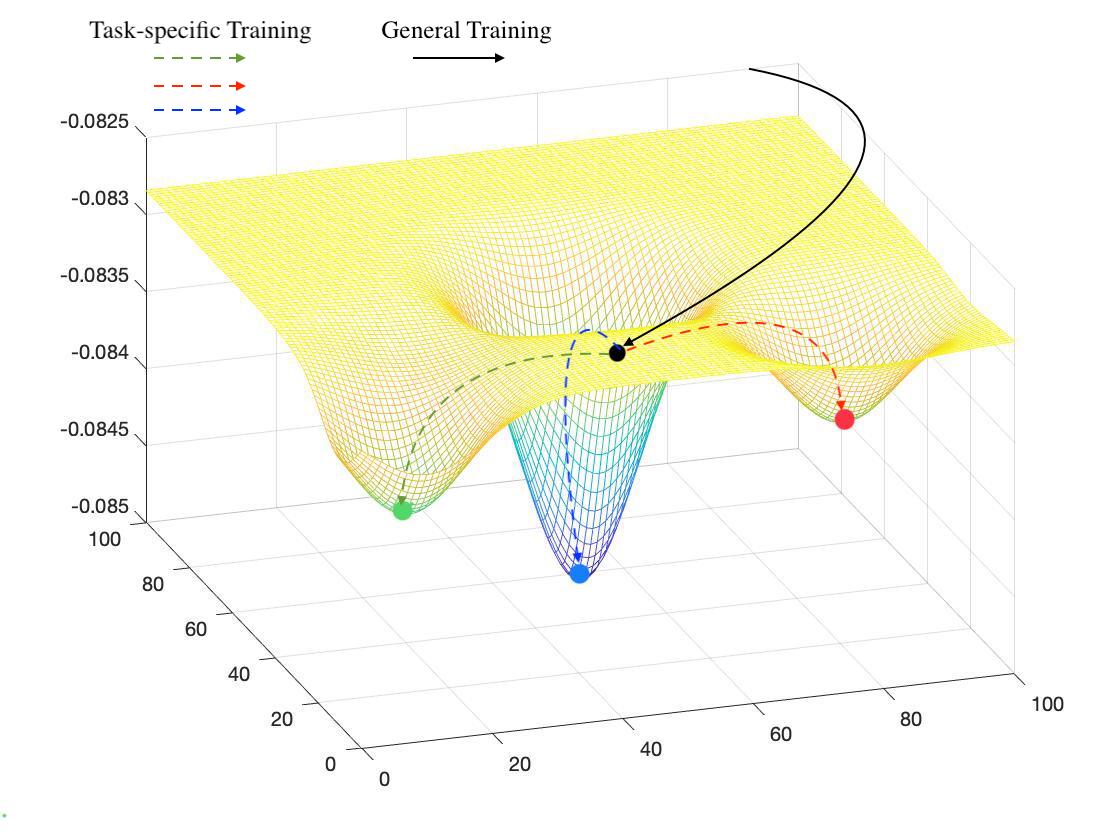}
    \caption{General Training and Task-specific Training}
    \label{fig:my_label}
\end{figure}

\subsection{Training Algorithm}
We divide our training algorithm into two parts - the general training part and the task-specific training part. General training searches for a global sub-optimal point shared by all tasks, and for each task, the followed task-specific training further helps the model to better fit the current situation.


\subsubsection{General Training Algorithm}
We can divide the general learning strategy into two steps - the first step applies the greedy approach to ensure the parameters of each task are progressing to the local optimized pattern and the second step takes into consideration all the local settings and strikes a balance between each location thus forwarding to the global optimal point(Figure 1).

\begin{algorithm}
\caption{General Training of SUM}
\begin{algorithmic}[1] 
\REQUIRE $\mathcal{D} = ( \mathcal{X} , \mathcal{Y} )$: data set over all tasks
\REQUIRE $\alpha , \beta, \lambda$:step size hyperparameters
\REQUIRE $p ( t )$: distribution of random time generation
\REQUIRE $S$: number of locations of the training set 
\STATE randomly initialize $\theta_{*}$
\WHILE{not converge}
\FOR{location i = 1:$S$ }
\STATE Randomly generate a time $t ^ { ( i ) } \sim p ( t )$
\STATE Evaluate $\nabla _{\theta_*}L(f(\mathcal{X}_{:,t^{(i)},i};\theta_*),\mathcal{Y}_{:,t^{(i)},i})$
\STATE Compute corresponding parameter with gradient descent:  $\theta_i = \theta_* - \alpha\nabla _{\theta_*}L(f(\mathcal{X}_{:,t^{(i)},i}; \theta_*),\mathcal{Y}_{:,t^{(i)},i})$
\ENDFOR
\STATE $\boldsymbol{W} = \left[vec(\theta_1),vec(\theta_2),...,vec(\theta_S)\right]$
\STATE  $\theta_* \xleftarrow{}\&\theta_* - \beta\nabla_{\theta_*}\sum_{i=1}^S\left[L(f(\mathcal{X}_{:,t^{(i)},i};\theta_i),\mathcal{Y}_{:,t^{(i)},i}) \&  +\lambda\cdot\text{tr}(\boldsymbol{W(D-A)W}^T) \right]]$
\ENDWHILE
\end{algorithmic}
\end{algorithm}

At first, we should ensure that the model we use for each task is structurally identical and is suitable for gradient descent. For each location, we randomly generate a time index $t^{(i)}$ from a probability distribution $p(t)$. $p(t)$ could be set differently according to different situations. Here we use a simple uniform distribution while we could also adjust it to have a heavy tail, which means the recent events matter more. 
For parameter updating, we first split the parameter from the current global point to each specific task by a step of gradient descent.
\[\theta_i = \theta_* - \alpha\nabla _{\theta_*}L(f(\mathcal{X}_{:,t^{(i)},i}; \theta_*),\mathcal{Y}_{:,t^{(i)},i})\eqno{(3)} \]

After renewing the local parameters for each location, we update the global setting according to the loss of all the tasks with a specially designed constraint.
\begin{align*}
\theta_* \xleftarrow{}&\theta_* - \beta\nabla_{\theta_*}\sum_{i=1}^S[L(f(\mathcal{X}_{:,t^{(i)},i};\theta_i),\mathcal{Y}_{:,t^{(i)},i}) \\&+ \lambda\cdot\text{tr}(\boldsymbol{W(D-A)W^T})\qquad\qquad\qquad{(4)}
\end{align*}
where the constraint $\lambda\cdot\text{tr}(\boldsymbol{W(D-A)W^T})$ is used to preserve the sptial relation between differnt tasks. In this constraint term, $\boldsymbol{A}$ is calculated by applying Gaussian kernel on the distances between each task.Specifically, let $d_{ij}$ denote the distance between task $i$ and task $j$, $\boldsymbol{A}_{ij} = exp[-d_{ij}/ \omega]$ and $\boldsymbol{D}$ is a diagonal matrix satisfying $\boldsymbol{D}_{ii} = \sum_j\boldsymbol{A}_{ij}$. $\lambda$ is used to balance the constraint and the gradient.

Another way to comprehend this step is that we fix the relative positions between the local parameters and the global parameters in hyperspace, and we optimize the global setting to minimize the losses of all tasks. In form, the sum of loss functions depends on the parameter of each set $\theta_i$ for i = 1,2,..,S. However, considering the formula 3, these parameters are in fact all dependent on $\theta_*$; thus making it feasible for derivation. 
Ideally,  $\theta_*$ will converge to the global optimal point.
\[\theta_* \xrightarrow{} \mathop{\arg\min}_{\theta_*}\sum_{i=1}^S L(f(\mathcal{X}_{:,:,i},\theta_i);\mathcal{Y}_{:,:,i})\eqno{(5)}\]

After such training, the algorithm will output a set of general parameters, which could be further fine-tuned on task-specific data.



\subsubsection{Task-specific Training}
For the setting of spatial-temporal prediction, data stream from a certain location may hold some special patterns along with the time series. To accommodate these task-specific features, we design task-specific training which allows further fine-tuning based on the pretrained parameters. This training procedure is detailed in Algorithm 2.


\begin{algorithm}
\caption{Task-specific Training of SUM}
\begin{algorithmic}[1] 
\REQUIRE $\mathcal{D}_{s} = ( \boldsymbol{X , Y} )$: data of one particular place
\REQUIRE $t_0$: the time of prediction
\REQUIRE $L$: the length of time for task-specific training
\REQUIRE $\theta_*$: parameters output by general training of SUM
\REQUIRE x: number of gradient descents for each time
\REQUIRE $\gamma$: step size parameter for only learning
\STATE $\theta = \theta_*$
\FOR{t = 1:L}
\FOR{i = 1:x}
\STATE Evaluate $\nabla_{\theta}L(f(X_{:,t}; \theta), Y_{:,t})$
\STATE $\theta \xleftarrow{} \theta - \gamma\nabla_{\theta}L(f(X_{:,t}; \theta), Y_{:,t}) $
\ENDFOR
\ENDFOR

\STATE Prediction $\hat{Y}_{:,t_0} = f(X_{:,t_0};\theta)$
\end{algorithmic}
\end{algorithm}

The number of updates and the step size parameter according to the temporal locality of the data could be flexibly selected under different situations. If the data has strong locality, then we could slightly increase the  number of updates and the step size to encourage faster learning on the task-specific data. Otherwise we could set smaller parameters to avoid over-fitting.

Besides, this algorithm does not require any geographical information; thus it is ideal for coKriging unknown places. The aim of this step is explicit: adjust the general pattern to meet a specific spatio-temporal condition and make a more accurate prediction for local data.

\section{Experiment}

We use the multi-scale spatio-temporal data to demonstrate the outstanding performance of the proposed learning framework. Specifically, we have enhanced the MUSCAT algorithm \cite{xu2018muscat} with SUM where the prediction is a linear combination of the latent variables on each scale. Detailed explanation of MUSCAT and SUM-MUSCAT are illustrated in the appendix.

\begin{table*}[htb]
\centering
\begin{tabular}{@{}p{4cm}p{6cm}p{6cm}p{6cm}p{6cm}@{}}
\toprule
           & \multicolumn{1}{c}{tmax} & \multicolumn{1}{c}{tmin} & \multicolumn{1}{c}{tmean} & \multicolumn{1}{c}{prcp} \\ \midrule
WISDOM     & \multicolumn{1}{c}{$0.3521 \pm 0.0165$}         & \multicolumn{1}{c}{$0.4023 \pm 0.0103$}        & \multicolumn{1}{c}{$0.3849 \pm 0.0198$}         & \multicolumn{1}{c}{$0.4123 \pm 0.0041$}        \\ \midrule
ALTO       & \multicolumn{1}{c}{$0.5923 \pm 0.0049$}        & \multicolumn{1}{c}{$0.5721 \pm 0.0018$}        & \multicolumn{1}{c}{$0.5642 \pm 0.0038$}         & \multicolumn{1}{c}{$0.5789 \pm 0.0038$}        \\ \midrule
MUSCAT     & \multicolumn{1}{c}{$0.3214 \pm 0.0041$}        & \multicolumn{1}{c}{$0.3465 \pm 0.0053$ }       & \multicolumn{1}{c}{$0.2846 \pm 0.0108$}         & \multicolumn{1}{c}{$0.4139 \pm 0.0034$}        \\ \midrule


SUM-MUSCAT & \multicolumn{1}{c}{$0.3120 \pm 0.0019$}        & \multicolumn{1}{c}{$0.3349 \pm 0.0108$}        &\multicolumn{1}{c}{$0.2802 \pm 0.0098$}         & \multicolumn{1}{c}{$0.4045 \pm 0.0049$}        \\ \bottomrule

\end{tabular}
\caption{MAE of Predictions}
\end{table*}

\subsection{Setting}
\subsubsection{Data}
The multi-scale spatio-temporal data is obtained from United States Historical Climatology Network(CSHCN)\footnote{http://cdiac.ornl.gov/epubs/ndp/ushcn/ushcn.html}, North American regional reanalysis dataset(NARR)\footnote{https://www.esrl.noaa.gov/psd/data/gridded/data.narr.html} and NCEP\footnote{http://www.esrl.noaa.gov/psd/data/gridded/data.ncep.reanalysis.derived.html}, which has a coarser resolution compared with NARR. We follow the same process of \cite{xu2018muscat} to construct the data set.

\subsubsection{Baseline Algorithm}
We compare SUM-MUSCAT with the following baselines, covering different multi-task learning approaches for spatial-temporal prediction.
\begin{enumerate}
    \item MUSCAT: The best performing spatial-temporal prediction algorithm proposed in \cite{xu2018muscat}.
    \item ALTO: The Low rank tensor learning methods proposed by \cite{yu2015accelerated} , which was designed to build models for multiple response variables simultaneously.
    \item WISDOM: The tensor decomposition method for spaital-temporal prection proposed in \cite{xu2016wisdom}.
\end{enumerate}

\subsection{Results}

\subsubsection{Climate Prediction}
Climate prediction aims to predict 4 climate features: the maximum(tmax), minimum(tmin) and mean(tmean) temparature and precipitation(prcp) of differnt locations. We followed the same settings used in \cite{xu2018muscat}.
As presented in Table 1, SUM-MUSCAT consistantly outperforms all the baselines including the current best performing model MUSCAT. Although MUSCAT has already achieved competitive performance with a rather complicated design, SUM could still improve its performance over 4 climate features.


\subsubsection{coKriging}

As for coKriging, many multi-task learning methods like ALTO\cite{yu2015accelerated} and WISDOM\cite{xu2016wisdom} are incapable of coKriging. Therefore, to demonstrate the performance of SUM-MUSCAT, we do comparisons with MTGP and MAML, two classical methods allowing coKriging.

We randomly selected 240 stations and randomly divided them into 24 groups. The former 80\% of the time period of the 240 stations are selected as training data while the latter 20\% are used for testing. We iteratively deleted one group(10 stations) in training set based on random decision and tested the performance on test set. We repeated 10 times of the above experiments and report the averaged result in Figure 2. 

We found that SUM has better generalization ability and achieves better results when do coKriging, which could be attributed to the proposed learning framework. Figure 2 also shows that, even with only 50\% of training data, our model could still achieve similar performance with using all the training data.

\begin{figure}[htp]
\centering
\includegraphics[width=4.2cm]{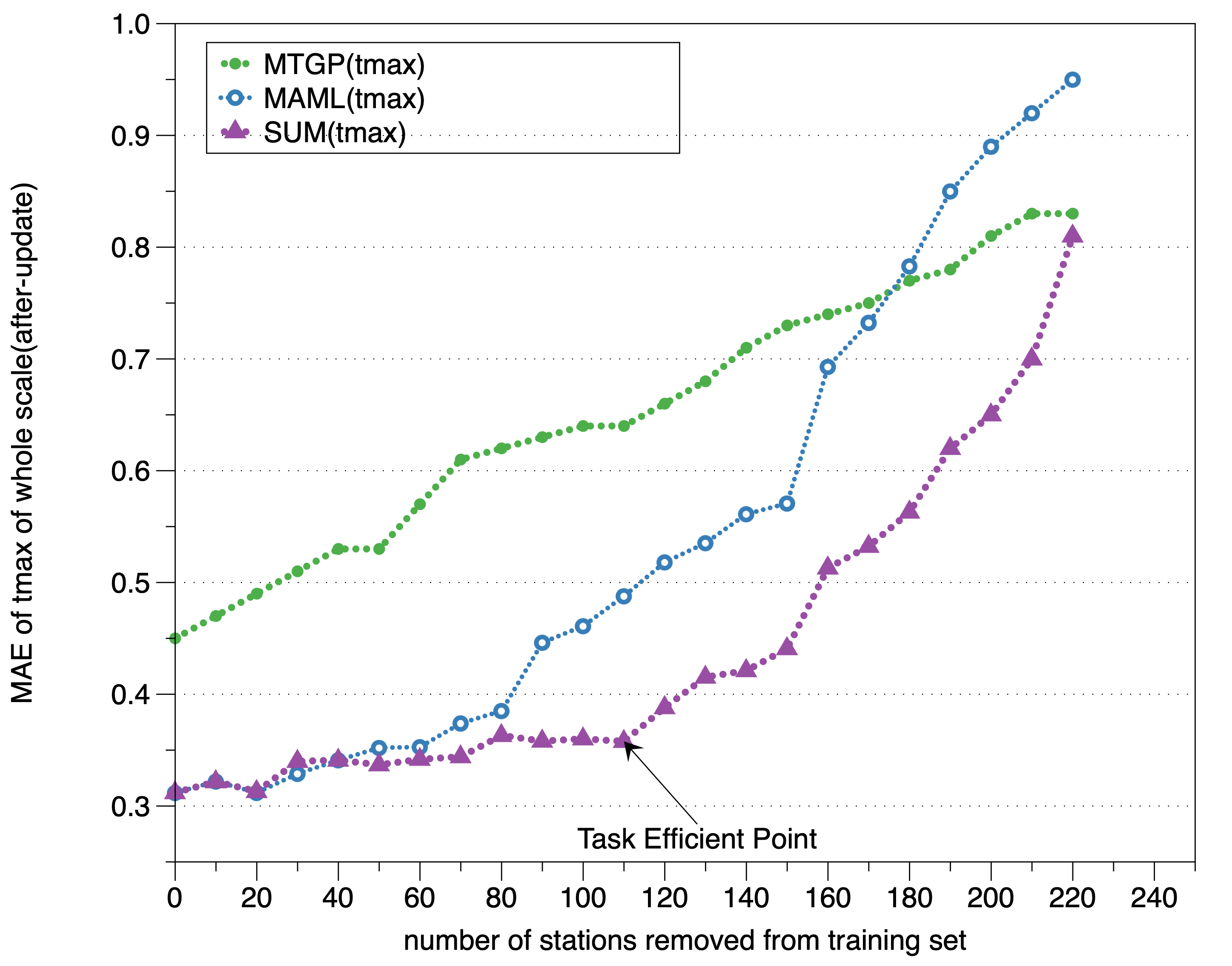}
\includegraphics[width=4.2cm]{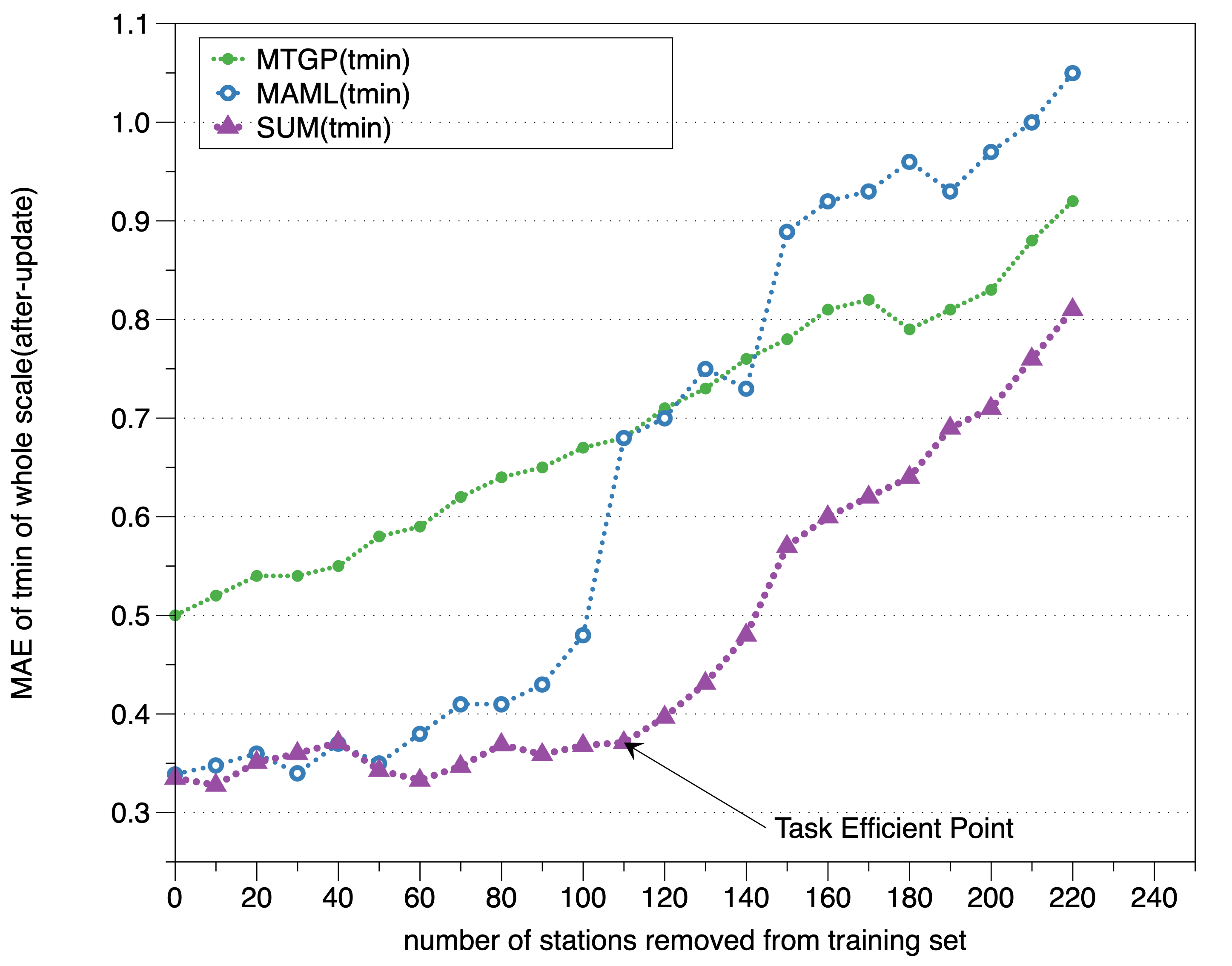}
\includegraphics[width=4.2cm]{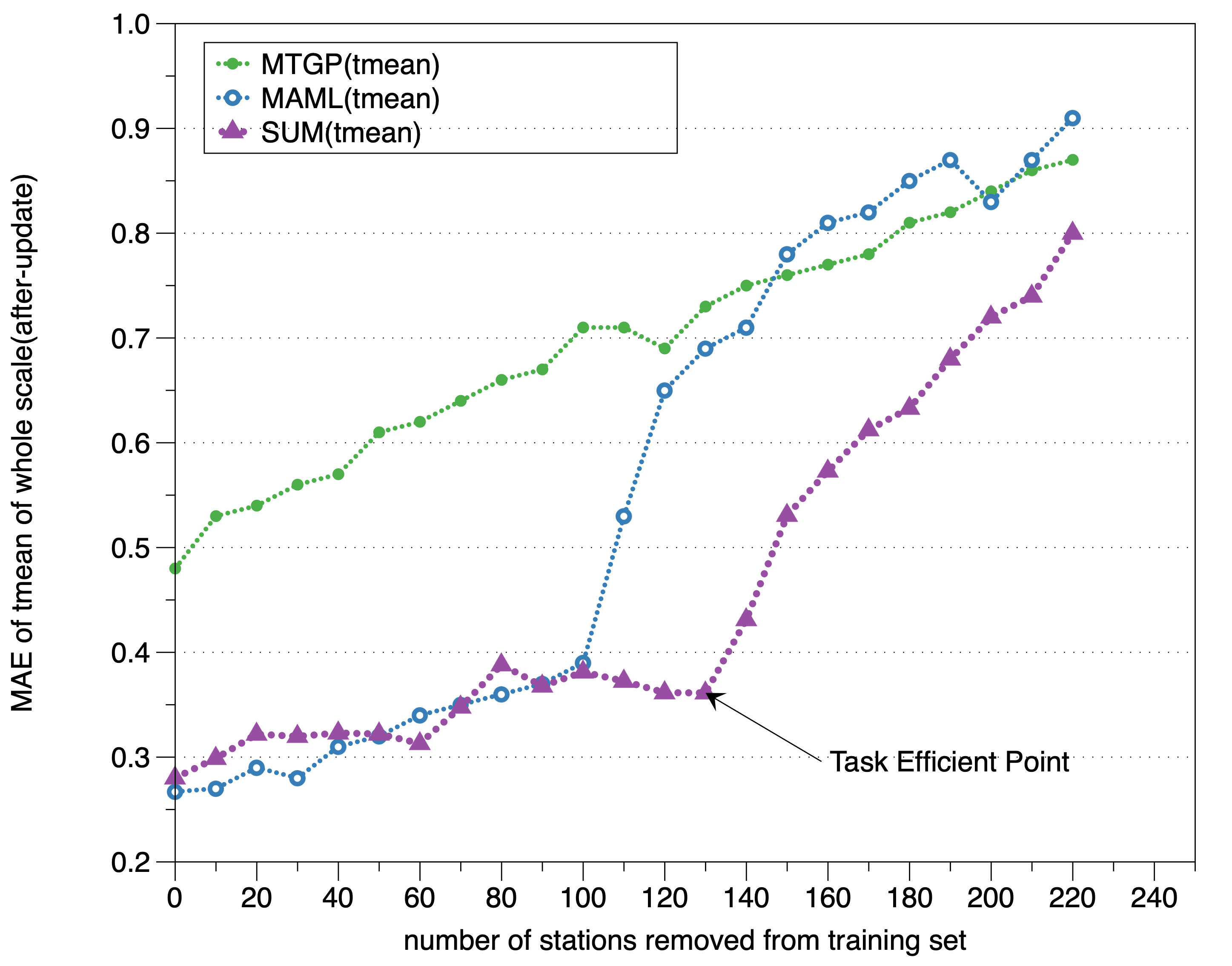}
\includegraphics[width=4.2cm]{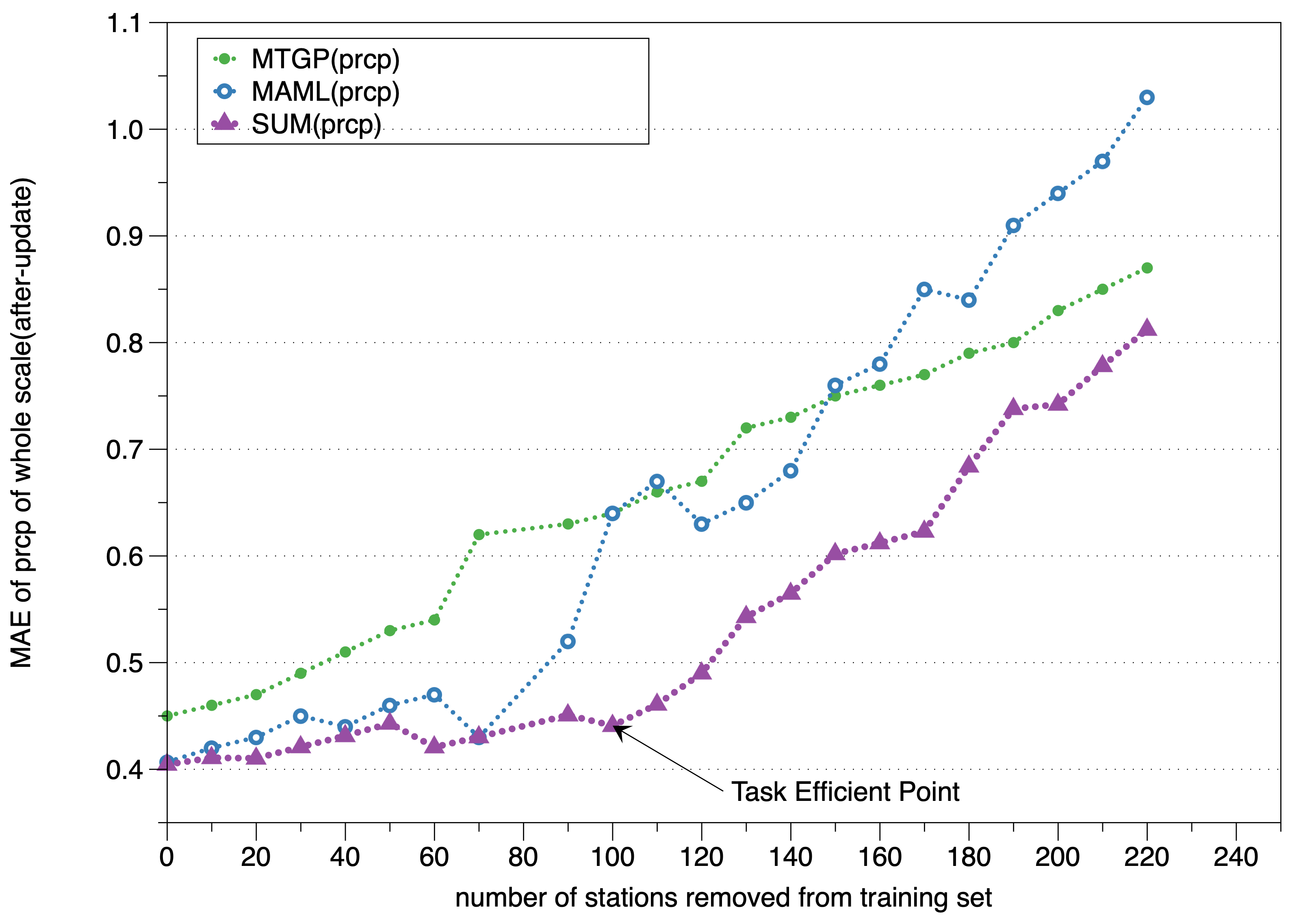}
\caption{MAE of coKriging}
\end{figure}

\subsubsection{Task Efficient Point}
Another intersting fact held in Figure 2 is about active learning. As we can see, there is a turning point from which the MAE starts to grow significantly compared with before, and we name this point as Task Efficient Point(TEP). This drives the following question that how is the training data distribution of our model differs from the others at TEP.
Figure 3 illustrated the task distribution of tmean at ETP of SUM-MUSCAT and MAML. We found that, with SUM, data at many stations is not needed for producing good performance. That is, the geographical information of different locations are better utilized by the proposed global learning strategy. 



\begin{figure}
    \centering
    \includegraphics[width=6cm]{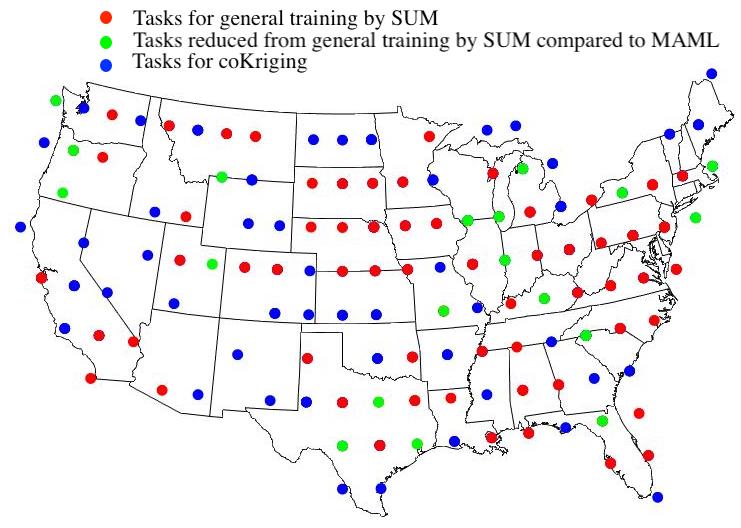}
    \caption{Task Distribution of tmean at Efficient Task Point}
    \label{fig:my_label}
\end{figure}

\section{Related Work}
How to extract useful information in spatial-temporal data and further predict the further trend has long been a hot topic in data mining. To handle the high dimensional problem of spatial-temporal data, low-rank tensor learning is used in many previous works. However, low-rank tensor learning is quite challenging due to the complexity of tensor analysis\cite{hillar2013most}. To address this problem, \cite{ouyang2013stochastic} \cite{avron2012efficient} use nuclear norm as a convex surrogate for the tank constraint. However, this treatment may lead to a sub-optimal solution and is computationally expensive. \cite{yu2015accelerated} devise an online learning algorithm and employs randomization techniques to overcome the issue of local optima. This method is extremely innovative but still cannot overcome the constraint underlining all low-rank tensor learning methods, which is the low quality of performance in a small batch. 

Tensor decomposition\cite{kolda2009tensor} is another way to extract essential information in spatio-temporal data streams, including precedent researches on coupled tensor decomposition \cite{signoretto2014learning}\cite{romera2013multilinear}. Many other works\cite{yu2015accelerated} \cite{liu2018distribution} involve the introduction of geographic information to guarantee the intercorrelation of each task. And we find it more than useful to ensure the generalization of multi-task models.

\section{Conclusion}
In this paper, we present a novel multi-task learning framework for spatial-temporal prediction. Benefiting from the two-step learning strategy, the proposed method demonstrates better performance and generalizability over existing methods.

SUM is also inclined with the recent fashion of pre-training and fine-tuning strategies in the area of natural language processing and computer vision. As a preliminary work which adopts such idea into spatial-temporal prediction, we hope this work could inspire further investigations in this direction.




\nocite{yu2015accelerated}
\nocite{bahadori2014fast}
\nocite{liu2018distribution}
\nocite{bonilla2008multi}
\nocite{finn2017model}
\nocite{caruana1997multitask}
\nocite{xu2016gspartan}
\nocite{xu2016wisdom}
\nocite{zhao2015multi}
\nocite{xu2018muscat}
\nocite{zhou2013tensor}
\nocite{romera2013multilinear}
\nocite{signoretto2014learning}

\bibliographystyle{named}
\bibliography{CIKM-19-Short}

\newpage
\newpage
\appendix
\label{app}
\section{Appendix}

\subsection{SUM-MUSCAT ----- a Specific Example}
In this section, we will provide an example of how we integrate SUM into  multi-scale spatio-temporal model and enable this model to make coKriging. Readers can consult this part to gain a deeper understanding of how to implement our framework to other models.
\subsubsection{MUSCAT Strategy}
At first, we use the method proposed in \cite{kolda2009tensor} to extract the invariant composition across different scales.
Different from the initial data set framework we proposed in the last chapter, we now have data set $\mathcal{D}=(\mathcal{X}^{(1)},...,\mathcal{X}^{(L)},Y)$where $\mathcal{X}^{(l)}$ denotes to the $l$-th scale of predictor variables.

We assume that we can decompose these predictor variables into underlying spatial and temporal factors. We use a tensor decomposition method CP\cite{kolda2009tensor} to extract these latent factors.
\[X ^ { ( L ) } = \llbracket{\boldsymbol{A , B , C ^ { ( l ) }}\rrbracket} = \sum _ { k = 1 } ^ { K } \boldsymbol{a} _ { k } \circ \boldsymbol{b} _ { k } \circ \boldsymbol{c} _ { k } ^ { ( l ) }\]
where $\boldsymbol{a}_k, \boldsymbol{b}_k, \boldsymbol{c}_k $ denotes to the k-th column vectors of matrix and $\circ$ denotes to the outer product operation. 

From the formula, we can see that $\boldsymbol{A} \in \mathbf{R}^{S \times K}, \boldsymbol{B} \in \mathbf{R}^{T \times K}$ , which denotes to the spatial and temporal latent factors, remain invariant across the scale. In our SUM model, we will degenerate these matrix parameters into vectors.

Furthermore, the response variables is a linear combination of each scale in the manner like
\[\hat { y } _ { t,s } = \sum _ { l } ^ { L } \alpha _ { l } \hat { y } _ { t , s } ^ { ( l ) }\]

and the $\alpha _ { l }$ denotes to the weight of the corresponding scale; thus satisfying $\sum _ { l } ^ { L } \alpha _ { l }=1$ and $\alpha _ { l } \geq 0$ for $l = 1,2,..L$.

As for $\hat { y } _ { s , t } ^ { ( l ) }$,  it is given by an ensemble of spatial and temporal models
\[\hat { y } _ { t , s } ^ { ( l ) } = \mathcal{X} _ {:, t , s } ^ { ( l ) ^ { T } } [ \sum _ { k } ^ { K } \boldsymbol{A} _ { s , k } \boldsymbol{w} _ { k } ^ { ( l ) } + \sum _ { k } ^ { K } \boldsymbol{B} _ { t , k } \boldsymbol{v} _ { k } ^ { ( l ) } ]\]
where $\boldsymbol{w} _ { k } ^ { ( l )}$ and $\boldsymbol{v} _ { k } ^ { ( l )}$ are parameters of the spatial and temporal prediction models for the k-th latent factor.

By simultaneously uncover the latent factors and derive the parameters for the spatial and temporal prediction, we can define lose function as follows:
\[\begin{split}L(\alpha,\boldsymbol{W},\boldsymbol{V},\boldsymbol{A},\boldsymbol{B},\boldsymbol{C})=  \| \sum _ { i } ^ { L } \alpha _ { i } \hat { \boldsymbol{Y} } ^ { ( l ) } - \boldsymbol{Y} \| _ { F } ^ { 2 } 
\\+ \frac { \lambda } { 2 } \sum _ { l } ^ { L } | | \mathcal{X} ^ { ( l ) } - \llbracket{\boldsymbol{A} , \boldsymbol{B} , \boldsymbol{C} ^ { ( l ) }\rrbracket}\| _ { F } ^ { 2 }\end{split}\]
which can be optimsed by incremental learning. 

For more detailed information, readers can consult the paper \cite{xu2018muscat}.

\subsubsection{SUM-MUSCAT Model}
Instead of considering different parameters for each time and location, we just  concerns the general pattern across all.  To do so, we degenerate the underlying spatial and temporal factors from matrix $\boldsymbol{A},\boldsymbol{B}$ into vectors $\boldsymbol{a},\boldsymbol{b}$. For each task, we only need to focus on the local features. Thus, we rewrite the tensor decomposition as follows:
\[
\mathcal{X} _ { : , t , s } ^ { ( L ) } = \llbracket{\boldsymbol{a} , \boldsymbol{b} , \boldsymbol{C} ^ { ( l ) }}\rrbracket = \sum _ { k = 1 } ^ { K } (a _ { k } \cdot b _ { k } )\circ \boldsymbol{c} _ { k } ^ { ( l ) }
\]

Therefore we can simply define the prediction function \[\hat { y } _ { t,s } = \sum _ { l } ^ { L } \alpha _ { l } \hat { y } _ { t , s } ^ { ( l ) }\]  where
\[\hat { y } _ { t , s } ^ { ( l ) } = \mathcal{X} _ {:, t , s} ^ { ( l ) ^ { T } } [ \sum _ { k } ^ { K } a_{k} \boldsymbol{w} _ { k } ^ { ( l ) } + \sum _ { k } ^ { K } b_{k}  \boldsymbol{v} _ { k } ^ { ( l ) } ]\]

Noticing that the original loss function is a non differentiable regarding $\sum _ { l } ^ { L } \alpha _ { l }=1$ and $\alpha _ { l } \geq 0$ for $l = 1,2,..L$, we can modify the lose function by taking the place of $\alpha_l$ by $k_l^2$ and adding a constraint to the lose function:

\begin{equation*}
\begin{split}
L(\hat{y}_{t,s}, y{t,s})= \frac{1}{2}(\hat{y}_{t,s}- y_{t,s})^2 + \frac{\lambda}{2}\sum_l^L\|\mathcal{X}_{:,t,s}^{(l)}-\llbracket{\boldsymbol{a},\boldsymbol{b}, \boldsymbol{C}^{(l)}}\rrbracket\|_2^2\\+\beta(1-\sum_l^L k_l^2)^2
\end{split}
\end{equation*}

And the prediction function will be

\begin{align*}
\hat { y } _ { t , s}
&= f ( \mathcal{X} _ { :,t , s } ^ { ( 1 ) },\mathcal{X} _ { :,t , s } ^ { ( 2 ) },...,\mathcal{X} _ { :,t , s } ^ { ( L ) } ; \boldsymbol{a} , \boldsymbol{b} , \boldsymbol{C} , \boldsymbol{k} , \boldsymbol{W} , \boldsymbol{V} )\\
&= \sum _ { l } ^ { L } k _ { l } ^ { 2 } \mathcal{X}_ {:, t , s } ^ { ( l ) ^ { T } }[ \sum _ { k } ^ { K } a _ { k } \boldsymbol{w} _ { k } ^ { ( l ) } + \sum _ { k } ^ { K } b _ { k } \boldsymbol{v} _ { k } ^ { ( l ) } ]
\end{align*}

This modification has somehow relaxed the primary constraints, but our experiment has proved that it can result in a better performance.

By implementing the above functions to algorithms proposed in the previous chapter, we can make both accurate and flexible predictions.

However, if the data given do not have the desired temporal locality, we can just degenerate the spatial factors, and the tensor decomposition should be as follows:
\[
\mathcal{X} _ { : , : , s } ^ { ( L ) } = \llbracket{\boldsymbol{a} , \boldsymbol{B} , \boldsymbol{C} ^ { ( l ) }}\rrbracket = \sum _ { k = 1 } ^ { K } (a _ { k } \cdot \boldsymbol{b} _ { k } )\circ \boldsymbol{c} _ { k } ^ { ( l ) }
\]
and the lose and prediction functions should have the corresponding change:
\[
\begin{split}
L(\hat{y}_{:,s}, y_{:,s})= \frac{1}{2}\|\hat{y}_{:,s}- y_{:,s}\|_2^2 + \frac{\lambda}{2}\sum_l^L\|\mathcal{X}_{:,:,s}^{(l)}-\llbracket{\boldsymbol{a},\boldsymbol{B}, \boldsymbol{C}^{(l)}}\rrbracket\|_2^2\\+\beta(1-\sum_l^L k_l^2)^2
\end{split}
\]

\begin{align*}
\hat { y } _ { t , s} 
&= f ( \mathcal{X} _ { :,t , s } ^ { ( 1 ) } ,\mathcal{X} _ { :,t , s } ^ { ( 2 ) },...\mathcal{X} _ { :,t , s } ^ { ( L ) }; \boldsymbol{a} , \boldsymbol{B} , \boldsymbol{C} , \boldsymbol{k} , \boldsymbol{W} , \boldsymbol{V} )\\
&= \sum _ { l } ^ { L } k _ { l } ^ { 2 } \mathcal{X}_ { :,t , s } ^ { ( l ) ^ { T } }[ \sum _ { k } ^ { K } a _ { k } \boldsymbol{w} _ { k } ^ { ( l ) } + \sum _ { k } ^ { K } \boldsymbol{B} _ { t,k } \boldsymbol{v} _ { k } ^ { ( l ) } ]
\end{align*}

Similarly, even the data has either strong time or spatial locality; we can still flexibly adjust our model to suit this circumstance by keeping both  $\boldsymbol{A,B} $ as its original matrix form, but doing so would impair the model's capability of coKriging.
We can see that the SUM model has added astonishing flexibility of parameters regarding $\boldsymbol{k},\boldsymbol{W},\boldsymbol{V}$ under all circumstances, as in the task-specific training, these parameters would be further optimised.

Moreover, compared with the original model, SUM-MUSCAT is capable of coKriging by implementing the strategy proposed in the previous section.

\end{document}